%% file: main.tex
\documentclass{article}

\usepackage[english]{babel}
\usepackage[numbers]{natbib}
\usepackage{float}
\usepackage{gensymb}
\usepackage[preprint]{neurips_2025}
\usepackage{setspace}
\usepackage{booktabs}
\usepackage{placeins}
\usepackage{tabularx}
\usepackage{makecell}
\usepackage{multirow}
\usepackage{amsmath}
\usepackage{graphicx}
\usepackage[skip=10pt, belowskip=10pt]{caption}
\usepackage[utf8]{inputenc} % allow utf-8 input
\usepackage[T1]{fontenc}    % use 8-bit T1 fonts
\usepackage{hyperref}       % hyperlinks
\usepackage{url}            % simple URL typesetting
\usepackage{amsfonts}       % blackboard math symbols
\usepackage{nicefrac}       % compact symbols for 1/2, etc.
\usepackage{microtype}      % microtypography
\usepackage{xcolor}         % colors
\usepackage{csquotes}

\title{8-Calves Image dataset}
\author{Xuyang Fang, Sion Hannuna, Neill Campbell, Edwin Simpson\\
University of Bristol\\
\{xf16910, sh1670, Neill.Campbell, edwin.simpson\}@bristol.ac.uk
}

\begin{document}
\maketitle

\input{0_Abstract/abstract}

\section{Introduction}

\input{1_Introduction/intro}

\section{Constructing 8-Calves dataset}

\input{2_Constructing_8-Calves_Dataset/dataset}

\section{Benchmarking object detectors on temporal calf detection}

\input{3_Transfomer_Detection/detection}

\section{Benchmarking vision models on temporal calf identification}

\input{4_Transfer_learning/transfer}

\section{Benchmarking Multi-Object Tracking}

\input{5_Tracking/track}

\section{Related works}

\input{6_Related_Work/related_work}

\section{Conclusion}

\input{7_Conclusion/conclusion}
\newpage
\bibliographystyle{IEEEtran}
\bibliography{reference}
\newpage
\appendix
\input{8_Appendix/appendix}

\end{document}

%% file: 0_Abstract/abstract.tex
\begin{abstract}
Automated monitoring of individual livestock is a cornerstone of 
precision livestock farming, but the development of robust computer 
vision models is hindered by a lack of datasets that capture the 
challenges of real-world group environments. To address this, we 
introduce the 8-Calves dataset, a challenging benchmark for 
multi-animal detection, tracking, and identification. The dataset 
features a one-hour video of eight Holstein Friesian calves 
in a barn, characterized by frequent occlusions, motion blur, 
and diverse poses. Through a semi-automated pipeline using a 
fine-tuned YOLOv8 detector and ByteTrack, followed by meticulous 
manual correction, we provide a high-quality resource of over 
537,000 bounding boxes with temporal identity labels.

We establish comprehensive baselines by benchmarking 28 
object detectors, revealing that while models achieve 
near-perfect performance on a lenient IoU threshold 
(mAP50: 95.2-98.9\%), their performance 
diverges on stricter metrics (mAP50:95: 56.5-66.4\%), highlighting 
the challenge of fine-grained localization. Our identification 
benchmark across 23 vision models demonstrates a 
trade-off: scaling model size improves classification accuracy 
but often compromises retrieval performance. We found that 
smaller, optimized architectures like ConvNextV2 Nano achieve 
the best balance (73.35\% accuracy, 50.82\% Top-1 KNN), 
outperforming larger variants and pure vision transformers. 
Furthermore, pre-training strategies focused on semantic 
learning (e.g., BEiT) yielded superior transferability. 
For multi-object tracking, our benchmarks reveal a significant 
gap: while leading trackers achieve high detection accuracy 
(MOTA > 0.92), they struggle severely with identity 
preservation (IDF1 $\approx$ 0.27), highlighting a key challenge 
for future research in occlusion-heavy scenarios.

The 8-Calves dataset bridges a gap in the literature 
by providing temporal richness, distinct identities, and 
realistic challenges. It serves as a resource 
for advancing vision models in agriculture, with benchmarks 
for detection, identification, and tracking, and opens 
avenues for future work in multimodal and self-supervised 
learning. The dataset and benchmark code are all publicly available at
https://huggingface.co/datasets/tonyFang04/8-calves.
\end{abstract}

%% file: 1_Introduction/intro.tex
\begin{figure}[t]
    \centering
    \includegraphics[width=0.90\textwidth]{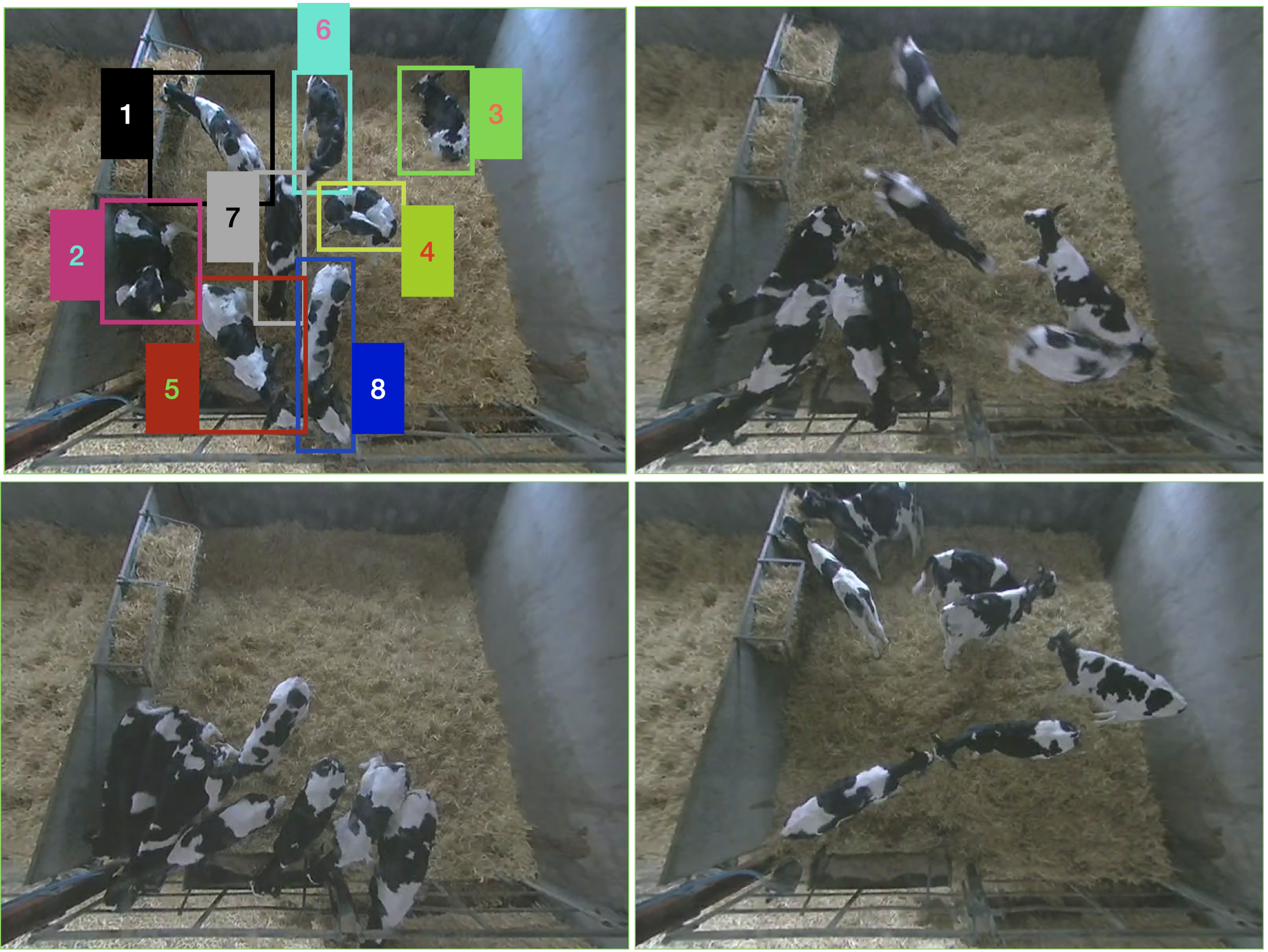}
    \caption{Representative frames from the 8-Calves dataset video: 
    (top-left) fully labeled frame; (top-right) frame with motion blur and 
    partial occlusion; (bottom-left) severe occlusion with indistinct calf 
    boundaries; (bottom-right) near-total occlusion as one calf is obscured 
    behind another.}
    \label{fig:data_screenshot}
\end{figure}

The monitoring of individual animals within a group is fundamental to 
advancing precision livestock farming (PLF). It enables early disease 
detection, optimizes feeding strategies, and improves our understanding 
of animal social behavior and welfare. Computer vision, with its 
potential for automated, non-invasive, and continuous monitoring, 
has emerged as a key technology in this domain. However, the development 
of robust vision models for such tasks is critically dependent on the 
availability of high-quality datasets that reflect the complex conditions 
of real-world agricultural environments.

While several datasets for cattle detection and identification exist, 
they often lack the temporal richness or the challenging conditions 
necessary to push the state of the art. Many benchmarks are composed 
of static images or short video clips, failing to capture the long-term 
temporal dynamics of group interactions. Others are captured in 
constrained settings that minimize occlusions and pose variations, 
or feature cattle breeds with low inter-animal visual distinguishability. 
However, the absence of benchmarks featuring long-term temporal continuity, 
visually distinct identities, and systematic occlusions makes it difficult 
to assess the reliability of models when confronted with these pervasive 
challenges in real deployments. Therefore, there 
is a clear need for a benchmark that combines temporal continuity, 
visually distinct identities, and dense, naturalistic occlusions to 
properly evaluate models for detection, tracking, and re-identification.

To address this gap, we introduce the 8-Calves dataset, a challenging 
new benchmark for multi-animal vision tasks. Our dataset comprises 
a video of eight Holstein Friesian calves in a barn 
during feeding, yielding over 537,000 high-quality, temporally 
consistent bounding boxes with unique identity labels. The scenario 
is characterized by frequent and sometimes severe occlusions, rapid motion, 
and diverse animal orientations, presenting a testbed for modern computer 
vision models, as illustrated in Figure~\ref{fig:data_screenshot}.

The primary contributions of this work are fourfold:

\begin{itemize}
    \item We present and describe the 8-Calves dataset, a long-range 
    temporal benchmark for calf detection, tracking, and identification, 
    collected in a challenging, occlusion-rich environment.
    \item We establish baselines by benchmarking a wide 
    array of state-of-the-art object detectors (28 models across YOLO 
    and Transformer-based families), revealing their performance 
    characteristics and limitations in fine-grained localization.
    \item We conduct an evaluation of 23 vision models 
    for calf identification, analyzing the impact of architecture, 
    model scale, and pre-training strategy on classification and 
    retrieval performance, and demonstrating a trade-off 
    between global and local feature learning.
    \item Our multi-object tracking evaluation demonstrates that 
    current methods, despite high detection scores, perform 
    poorly on identity preservation, underscoring the 
    challenge posed by our dataset.
\end{itemize}

By making this dataset and our benchmarking results publicly 
available, we aim to foster progress in the development of robust 
vision systems for animal monitoring. The 8-Calves dataset serves 
not only as a challenging benchmark but also as a 
resource for exploring future research in multimodal learning and 
self-supervised training. Additionally, this capability to reliably 
track and identify individuals has the potential to contribute to 
the automation of key management tasks. For instance, the temporal 
identity labels can be directly utilized to:
\begin{itemize}
    \item Monitor feeding behavior: By analyzing an individual's 
    presence at the feeding rack, we can automatically calculate 
    feeding duration and frequency, enabling the optimization of 
    feed distribution and early detection of anorexia---a common
    sign of illness.
    \item Quantify social behavior: Tracking spatial proximity and 
    interactions between identified calves allows for the assessment 
    of social hierarchies and cohesion, which are critical indicators 
    of animal welfare.
    \item Enable early disease detection: Subtle changes in gait or 
    posture indicative of lameness can be identified by analyzing 
    the movement patterns of specific individuals over time.
\end{itemize}

Therefore, by providing a high-quality benchmark for the underlying 
computer vision tasks, this work can facilitate the development of 
intelligent systems to address practical agricultural challenges.

The remainder of this paper is structured as follows: Section 2 
details the creation and characteristics of the 8-Calves dataset. 
Sections 3, 4, and 5 present our benchmarking results for object 
detection, calf identification, and multi-object tracking, 
respectively. To ensure the reliability and statistical significance 
of our findings, all experiments in this work are repeated 
three times, and the mean and standard deviation of each are 
reported. Section 6 discusses related works, and Section 7 
concludes the paper and outlines future directions.

%% file: 2_Constructing_8-Calves_Dataset/dataset.tex
Our raw data comprises nine one-hour videos featuring the same eight 
Holstein Friesian calves in a barn during feeding periods, with a resolution 
of \(600\times 800\) pixels and a frame rate of 20 fps. The setting involves frequent 
occlusions, varied poses, orientations, aspect ratios, and occasional motion 
blur due to rapid movement. To ensure high-quality ground-truth bounding boxes
and identities, we focused our manual efforts on a single, representative one-hour
video characterized by dense animal interactions during feeding.

We used a semi-automated approach to minimize manual labeling of calf bounding boxes 
and identities. ByteTrack \cite{bytetrack}, a top-performing multi-object tracker,
was applied to the video, followed by manual error correction. Since 
ByteTrack \cite{bytetrack} requires an object detector, we fine-tuned a YOLOv8m \cite{ultralytics_yolov8}
model from Ultralytics. The training data consisted of 900 randomly selected 
frames (100 from each of the nine original videos), which were manually annotated.

After integrating the detector with ByteTrack \cite{bytetrack} and processing all footage, 
we performed full error correction (false positives, identity swaps, missed tracks) 
by hand. The final dataset contains 537,908 validated bounding boxes and identities. 
Manual inspection revealed \(\sim 3,000\) missing detections (0.56\% of total), which 
negligibly impacts detector training/evaluation and does not affect identity tracking.

%% file: 3_Transfomer_Detection/detection.tex
\subsection{Experimental Setup}

We evaluated state-of-the-art detectors by fine-tuning them on a limited subset of our 
dataset. For comprehensive testing, all 67,760 labeled video frames were used as the 
test set. To prevent data leakage, only 600 out of 900 randomly selected and manually 
labeled frames were allocated for training and validation. These 600 frames were 
arranged in chronological order, with the first 500 used for training and the 
remaining 100 for validation---the latter also serving to determine early stopping. 

We selected 28 models for benchmarking, including multiple variants of YOLO series 
(YOLOv6/v8/v10/v11: n, s, m, l, x; YOLOv9: t, s, e, c, m 
\cite{yolov6,ultralytics_yolov8,yolov9,yolov10,yolo11_ultralytics}), along with 
transformer-based detectors such as Facebook DETR-ResNet50 \cite{carion2020detr}, 
Conditional DETR-ResNet50 \cite{conditionalDETR}, 
and YOLOS-Small \cite{YOLOS} (which employs a small Google Vision Transformer 
as backbone and is not a YOLO detector). All models were pre-trained on 
the MS-COCO dataset.

Due to GPU memory constraints, we set the batch size to 5 for all models and excluded 
larger architectures such as Deformable DETR \cite{deformableDETR} and DETR-ResNet101 
\cite{carion2020detr}, focusing instead on models deployable on mid-tier hardware. 
To reduce training time, we incorporated an early stopping mechanism. The input data 
were not normalized using ImageNet-1K \cite{ImageNet} statistics, as their distribution 
differs significantly from that of our dataset. All other hyperparameters were kept 
consistent with the original configurations provided in each model's respective paper. 
Further hyperparameter details can be found in Appendix A, Table~\ref{tab:hyperparams}.

During training, we applied the following data augmentations: 90\textdegree\space
rotation, horizontal flip, brightness-contrast adjustment, and elastic transformation. 
We deliberately avoided augmentations that could alter the fixed number of cows 
per image or shift the underlying data distribution---such as copy-paste, mosaic, 
perspective warping, mixup, cropping, erasing, and grayscale conversion. Due to 
time constraints, the augmentation probabilities were fixed across all models: 
1.0 for 90\textdegree\space rotation, 0.5 for horizontal flip, 0.4 for 
brightness-contrast adjustment, and 0.5 for elastic transformation 
(with \(\alpha = 100.0\) and \(\sigma = 5.0\)).

\subsection{Results and Analysis}

\begin{table}[htpb]

\caption{Object detection performance on the 8-Calves dataset. 
The best, second, and third performers are highlighted in red, blue, 
and green, respectively.}
\vspace*{10pt} % 10pt vertical space after the caption
\centering
\label{tab:object_detector_performance_table}
\begin{tabular}{lcccc}
\toprule
\textbf{Model} & \makecell{\textbf{Parameters} \\ \textbf{(M)}} & \makecell{\textbf{mAP50:95} \\ \textbf{(\%)}} & \makecell{\textbf{mAP75} \\ \textbf{(\%)}} & \makecell{\textbf{mAP50} \\ \textbf{(\%)}} \\
\midrule
% YOLOv6
yolov6n  & 4.5   & 61.3 \(\pm\) 0.3 & 70.5 \(\pm\) 0.3 & 97.8 \(\pm\) 0.4 \\
yolov6s  & 16.4  & 62.6 \(\pm\) 1.0 & 73.0 \(\pm\) 2.1 & 98.2 \(\pm\) 0.1 \\
yolov6m  & 52.0  & \textcolor{red}{66.4 \(\pm\) 1.3} & \textcolor{red}{78.9 \(\pm\) 2.5} & \textcolor{blue}{98.8 \(\pm\) 0.1} \\
yolov6l  & 110.9 & 63.6 \(\pm\) 1.0 & 75.1 \(\pm\) 1.5 & 98.0 \(\pm\) 0.3 \\
yolov6x  & 173.1 & 62.3 \(\pm\) 0.4 & 72.8 \(\pm\) 0.7 & 97.2 \(\pm\) 0.6 \\
\midrule
% YOLOv8
yolov8n  & 3.2   & 61.5 \(\pm\) 2.6 & 71.0 \(\pm\) 5.0 & 97.6 \(\pm\) 1.2 \\
yolov8s  & 11.2  & 62.5 \(\pm\) 2.6 & 72.4 \(\pm\) 4.7 & 98.0 \(\pm\) 0.8 \\
yolov8m  & 25.9  & 64.3 \(\pm\) 0.7 & 75.5 \(\pm\) 1.7 & 98.5 \(\pm\) 0.5 \\
yolov8l  & 43.7  & 64.8 \(\pm\) 2.4 & 76.6 \(\pm\) 3.7 & 98.6 \(\pm\) 0.7 \\
yolov8x  & 68.2  & \textcolor{blue}{65.7 \(\pm\) 2.2} & \textcolor{blue}{77.8 \(\pm\) 3.5} & \textcolor{green}{98.8 \(\pm\) 0.3} \\
\midrule
% YOLOv9
yolov9t  & 2.1   & 62.5 \(\pm\) 2.0 & 72.7 \(\pm\) 4.1 & 98.3 \(\pm\) 0.5 \\
yolov9s  & 7.3   & 64.7 \(\pm\) 1.1 & 76.4 \(\pm\) 1.7 & 98.8 \(\pm\) 0.2 \\
yolov9m  & 20.2  & 63.4 \(\pm\) 1.1 & 74.0 \(\pm\) 2.0 & 98.4 \(\pm\) 0.7 \\
yolov9c  & 25.6  & \textcolor{green}{64.9 \(\pm\) 3.1} & \textcolor{green}{77.0 \(\pm\) 5.1} & 98.6 \(\pm\) 0.5 \\
yolov9e  & 58.2  & 64.5 \(\pm\) 0.9 & 76.1 \(\pm\) 1.9 & \textcolor{red}{98.9 \(\pm\) 0.1} \\
\midrule
% YOLOv10
yolov10l & 25.9  & 62.2 \(\pm\) 1.4 & 72.2 \(\pm\) 2.6 & 96.9 \(\pm\) 0.6 \\
yolov10m & 16.6  & 61.8 \(\pm\) 1.2 & 71.6 \(\pm\) 2.5 & 97.0 \(\pm\) 0.4 \\
yolov10n & 2.8   & 59.9 \(\pm\) 4.1 & 68.1 \(\pm\) 7.3 & 96.1 \(\pm\) 1.7 \\
yolov10s & 8.1   & 61.2 \(\pm\) 3.6 & 70.7 \(\pm\) 6.5 & 96.5 \(\pm\) 1.8 \\
yolov10x & 31.8  & 60.2 \(\pm\) 5.3 & 69.0 \(\pm\) 9.3 & 95.2 \(\pm\) 3.0 \\
\midrule
% YOLOv11
yolo11l & 25.4  & 62.1 \(\pm\) 0.3 & 71.6 \(\pm\) 0.7 & 98.6 \(\pm\) 0.0 \\
yolo11m & 20.1  & 62.3 \(\pm\) 3.3 & 72.0 \(\pm\) 5.1 & 98.0 \(\pm\) 0.8 \\
yolo11n & 2.6   & 61.0 \(\pm\) 2.7 & 70.0 \(\pm\) 5.6 & 97.7 \(\pm\) 0.8 \\
yolo11s & 9.5   & 61.5 \(\pm\) 4.5 & 70.8 \(\pm\) 7.7 & 97.4 \(\pm\) 1.3 \\
yolo11x & 57.0  & 64.2 \(\pm\) 1.0 & 75.7 \(\pm\) 0.9 & 98.5 \(\pm\) 0.4 \\
\midrule
\midrule
Facebook DETR ResNet50    & 41.5 & 60.0 \(\pm\) 0.2 & 68.3 \(\pm\) 0.4 & 97.2 \(\pm\) 0.1 \\
YOLOS Small               & 30.7 & 56.5 \(\pm\) 0.7 & 61.6 \(\pm\) 1.5 & 95.9 \(\pm\) 0.6 \\
Conditional DETR ResNet50 & 43.4 & 61.0 \(\pm\) 1.3 & 69.8 \(\pm\) 2.8 & 98.0 \(\pm\) 0.4 \\
\bottomrule
\end{tabular}
\end{table}

Table\ref{tab:object_detector_performance_table} summarizes the benchmarking 
results of all evaluated models. While each 
model achieved high mAP50 scores ranging from 95.2\% to 98.9\%, indicating 
near-ceiling performance under a lenient IoU threshold, a significant performance 
drop was observed in mAP75 (61.6--78.9\%) and mAP50:95 (56.5--66.4\%). This 
decline highlights the challenge of fine-grained detection posed by our dataset. 
Moreover, the wider performance spread in these stricter metrics demonstrates 
the dataset's ability to reveal meaningful disparities between different model 
architectures.

The top-performing models in fine-grained detection were YOLOv6m (66.4\% 
mAP50:95, 78.9\% mAP75) and YOLOv8x (65.7\% mAP50:95, 77.8\% mAP75), 
followed by YOLOv9c (64.9\% mAP50:95, 77.0\% mAP75). YOLOv9t is the most 
efficient model, containing only 2.1 million parameters---the fewest among 
all variants. Despite its compact size, it achieves the best 
parameter-to-performance ratio, making it highly suitable for edge 
deployment. Comparisons among transformer-based models reveal the 
limitations of pure vision transformer architectures (e.g., YOLOS-Small), 
while underscoring the effectiveness of hybrid designs such as 
Conditional DETR and Facebook DETR.

\begin{figure}[htpb]
    \centering
    \includegraphics[width=0.95\textwidth]{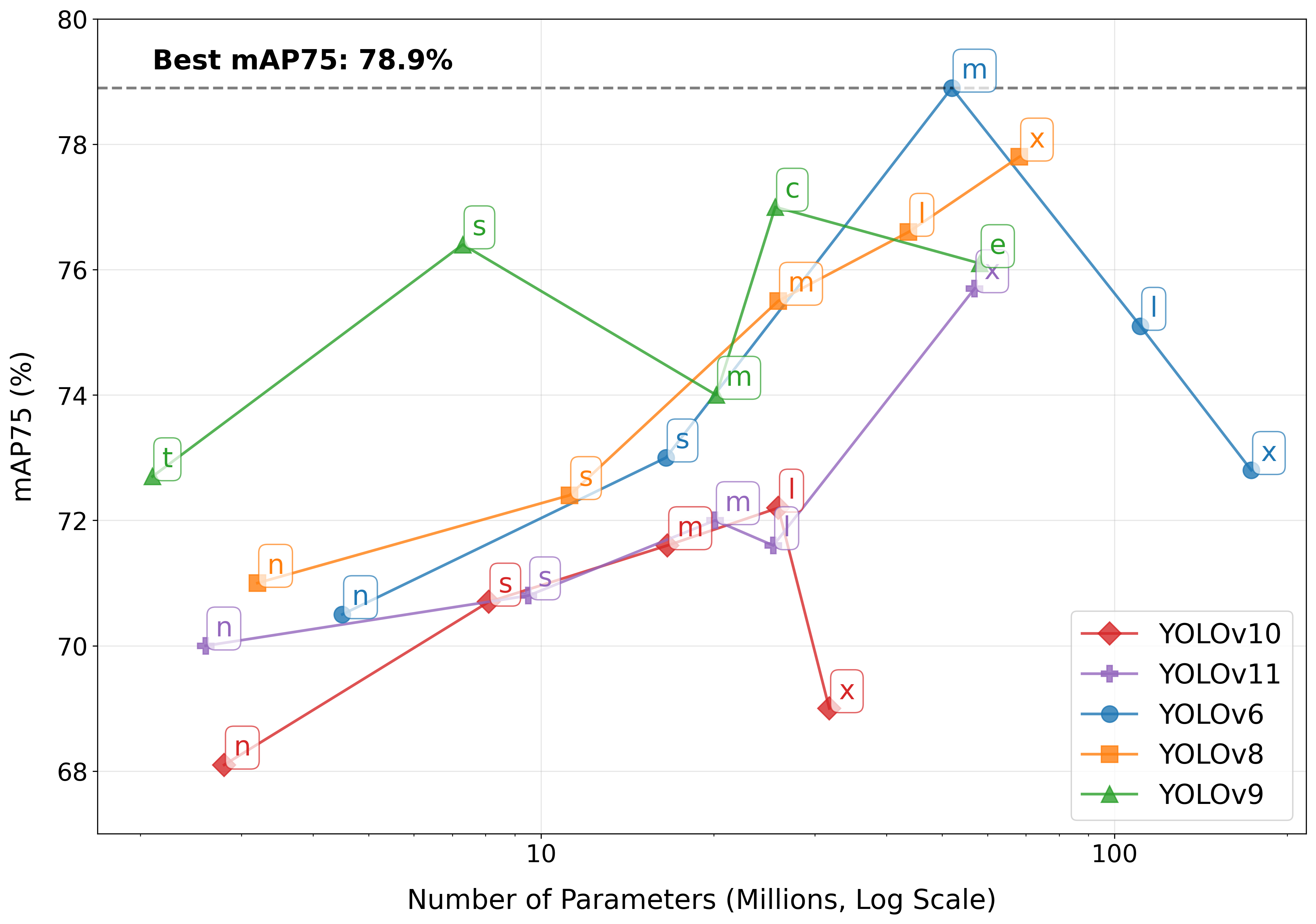}
    \caption{Benchmarking results of YOLO series on the 8-Calves dataset.}
    \label{fig:map75_vs_num_params_plot}
\end{figure}

The YOLOv6 and YOLOv10 series both exhibit signs of overfitting as model 
size increases exponentially. Within the YOLOv6 family, 
while the medium variant (YOLOv6m, 52M parameters) achieves the highest 
mAP50:95 of 66.4\%, larger models such as YOLOv6l (110.9M parameters, 
63.6\%) and YOLOv6x (173.1M parameters, 62.3\%) show a clear decline 
in performance. A similar trend is observed in the YOLOv10 series, 
where the extra-large variant (YOLOv10x, 31.8M parameters, 60.2\%) 
underperforms compared to the large variant (YOLOv10l, 25.9M parameters, 62.2\%).
These performance trends across the YOLO families are visualized in Figure~\ref{fig:map75_vs_num_params_plot}.

The presence of \(\sim 3,000\) false negatives introduces a maximum uncertainty of 
\(\pm 0.5\%\) in the reported mAP values. This uncertainty arises because mAP 
is calculated as the area under the precision--recall curve, which is influenced 
by the number of undetected ground-truth instances. If all 3,000 false negatives 
remained undetected, mAP would be overestimated by approximately 0.5\%. 
Conversely, if all were correctly detected with the highest confidence among 
predictions, mAP would be underestimated by a similar margin. It should be noted 
that although the labeling results underwent human verification, the test 
set may still exhibit bias toward YOLO-based detectors, as YOLOv8m 
\cite{ultralytics_yolov8} was used for initial annotation. Therefore, 
direct comparison between YOLO detectors and transformer-based models using mAP 
metrics should be interpreted with caution.

%% file: 4_Transfer_learning/transfer.tex
\subsection{Experimental Setup}
Our dataset comprises 537,908 ground truth bounding boxes with unique 
identifiers, enabling a robust evaluation of transfer learning 
performance across various vision models. We began by selecting 
23 sets of model weights, each pretrained on \(224 \times 224\) pixel 
images. To ensure a fair comparison, the models were divided into 
two groups. Group 1 includes four architectures: ResNet \cite{ResNet}, 
ConvNextV2 \cite{ConvNextV2}, Swin Transformers \cite{SwinTransformer}, 
and the original Vision Transformers (ViT) from Google \cite{ViT}, 
all initially pretrained on ImageNet \cite{ImageNet}. However, it 
should be noted that the comparison within this group is biased 
against ResNet models \cite{ResNet}, as they were pretrained solely 
on ImageNet-1K \cite{ImageNet}, whereas the others were pretrained 
on ImageNet-22K \cite{ImageNet} before being fine-tuned on ImageNet-1K 
\cite{ImageNet}.

Group 2 included six model weights, all built upon the ViT-Base 
backbone \cite{ViT} but differing in their pretraining strategies. 
The self-supervised learning models comprised DinoV2 \cite{DinoV2} 
(pretrained on LVD-142M), MAE \cite{Vit-MAE}, and BEiT \cite{BEIT} 
(both pretrained on ImageNet-1K \cite{ImageNet}). In contrast, 
CLIP \cite{CLIP} adopted a multimodal approach using 400 million 
image-text pairs. Lastly, DeiT \cite{DEIT} was trained on ImageNet-1K
\cite{ImageNet} with knowledge distillation from a convolutional 
neural network teacher model.

We extracted all bounding boxes from the video data and resized 
them to \(224 \times 224\) pixels. These image patches were then 
forward-passed through all 23 models without standardization to 
ImageNet's mean \cite{ImageNet}, resulting in 23 distinct embedding 
spaces. These embeddings were subsequently used in two downstream 
tasks: Temporal Linear Classification and Temporal K-Nearest 
Neighbors (KNN). The former assessed the global structure of the 
embedding space---specifically, whether a linear classifier could 
effectively distinguish among the eight calves. The latter evaluated 
the local geometry, determining whether embeddings from the same 
calf clustered closely together.

For both tasks, the embedding spaces were split chronologically. 
In the Temporal Linear Classification task, we employed a 30/30/40 
split for training, validation, and testing, respectively. The 
validation set was used for hyperparameter tuning and early stopping. 
The classifier was trained for up to 100 epochs with a batch size of 
32, using the AdamW optimizer (learning rate = 0.005), cross-entropy 
loss, and variance scaling for parameter initialization. Early stopping 
was triggered if no improvement was observed for 10 epochs. For the 
Temporal KNN task, a simpler 50/50 train-test split was applied. To 
accelerate KNN inference, we utilized FAISS with an IndexIVFPQ quantizer, 
configured with 8 subquantizers, 8 bits per subquantizer, and 10 probes.

\begin{table}[htpb!]
\centering
\caption{Calf identification performance on the 8-Calves dataset. For each architecture family, the best-performing variant is highlighted in red.}
\vspace*{10pt} % 10pt vertical space after the caption
\label{tab:transfer_learning_model_family_comparison}
\begin{tabular}{cccccc}
\toprule
\makecell{\textbf{Architecture} \\ \textbf{Family}} & \makecell{\textbf{Model} \\ \textbf{Variant}} & \makecell{\textbf{Accuracy} \\ \textbf{(\%)}} & \makecell{\textbf{KNN} \\ \textbf{Top-1 (\%)}} & \makecell{\textbf{KNN} \\ \textbf{Top-5 (\%)}} & \textbf{Parameters} \\
\midrule
\multirow{7}{*}{\shortstack{ConvNextV2\\[0.1em]Patch Size=4}} 
  & Atto & 65.79 \(\pm\) 0.09 & 48.08 \(\pm\) 0.20 & 70.71 \(\pm\) 0.19 & 3.7M \\
  & Pico & 72.48 \(\pm\) 0.60 & 36.81 \(\pm\) 0.00 & 63.01 \(\pm\) 0.00 & 9.1M \\
  & \textbf{Nano} & \textcolor{red}{\textbf{73.35 \(\pm\) 0.27}} & \textcolor{red}{\textbf{50.82 \(\pm\) 0.00}} & \textcolor{red}{\textbf{71.63 \(\pm\) 0.00}} & 15.6M \\
  & Tiny & 72.61 \(\pm\) 0.05 & 46.42 \(\pm\) 0.14 & 70.66 \(\pm\) 0.12 & 28.6M \\
  & Base & 66.40 \(\pm\) 0.02 & 41.13 \(\pm\) 0.08 & 66.10 \(\pm\) 0.02 & 89.0M \\
  & Large & 67.65 \(\pm\) 0.48 & 37.65 \(\pm\) 0.01 & 65.31 \(\pm\) 0.03 & 196.4M \\
  & Huge & 42.28 \(\pm\) 1.31 & 27.52 \(\pm\) 0.04 & 54.29 \(\pm\) 0.14 & 657.5M \\
\midrule
\multirow{4}{*}{\shortstack{Swin Transformer\\[0.1em]Patch Size=4}} 
  & Tiny & 68.39 \(\pm\) 0.31 & \textcolor{red}{\textbf{43.87 \(\pm\) 0.01}} & \textcolor{red}{\textbf{67.76 \(\pm\) 0.04}} & 28.3M \\
  & Small & 65.99 \(\pm\) 0.15 & 38.73 \(\pm\) 0.15 & 63.29 \(\pm\) 0.01 & 50.0M \\
  & Base & 67.34 \(\pm\) 0.01 & 36.47 \(\pm\) 0.00 & 63.52 \(\pm\) 0.00 & 87.8M \\
  & Large & \textcolor{red}{\textbf{70.47 \(\pm\) 0.17}} & 35.76 \(\pm\) 0.00 & 61.31 \(\pm\) 0.00 & 197.0M \\
\midrule
\multirow{5}{*}{ResNet} 
  & 18 & 60.45 \(\pm\) 0.00 & \textcolor{red}{\textbf{45.41 \(\pm\) 0.02}} & \textcolor{red}{\textbf{65.42 \(\pm\) 0.15}} & 11.7M \\
  & 34 & 56.00 \(\pm\) 0.34 & 37.24 \(\pm\) 0.10 & 59.99 \(\pm\) 0.25 & 21.8M \\
  & 50 & 62.21 \(\pm\) 1.53 & 38.60 \(\pm\) 0.35 & 59.53 \(\pm\) 0.30 & 25.6M \\
  & 101 & \textcolor{red}{\textbf{64.56 \(\pm\) 1.08}} & 37.22 \(\pm\) 0.02 & 57.62 \(\pm\) 0.18 & 44.5M \\
  & 152 & 62.45 \(\pm\) 0.18 & 36.71 \(\pm\) 0.02 & 58.46 \(\pm\) 0.12 & 60.2M \\
\midrule
\multirow{2}{*}{\shortstack{ViT\\[0.1em]Patch Size=16}} 
  & Base & 54.70 \(\pm\) 0.42 & \textcolor{red}{\textbf{29.53 \(\pm\) 0.25}} & \textcolor{red}{\textbf{52.90 \(\pm\) 0.34}} & 86.4M \\
  & Large & \textcolor{red}{\textbf{56.12 \(\pm\) 0.52}} & 28.15 \(\pm\) 0.06 & 51.35 \(\pm\) 0.18 & 304.4M \\
\bottomrule
\end{tabular}
\end{table}

\begin{figure}[!htbp]
    \centering
    \includegraphics[width=0.95\textwidth]{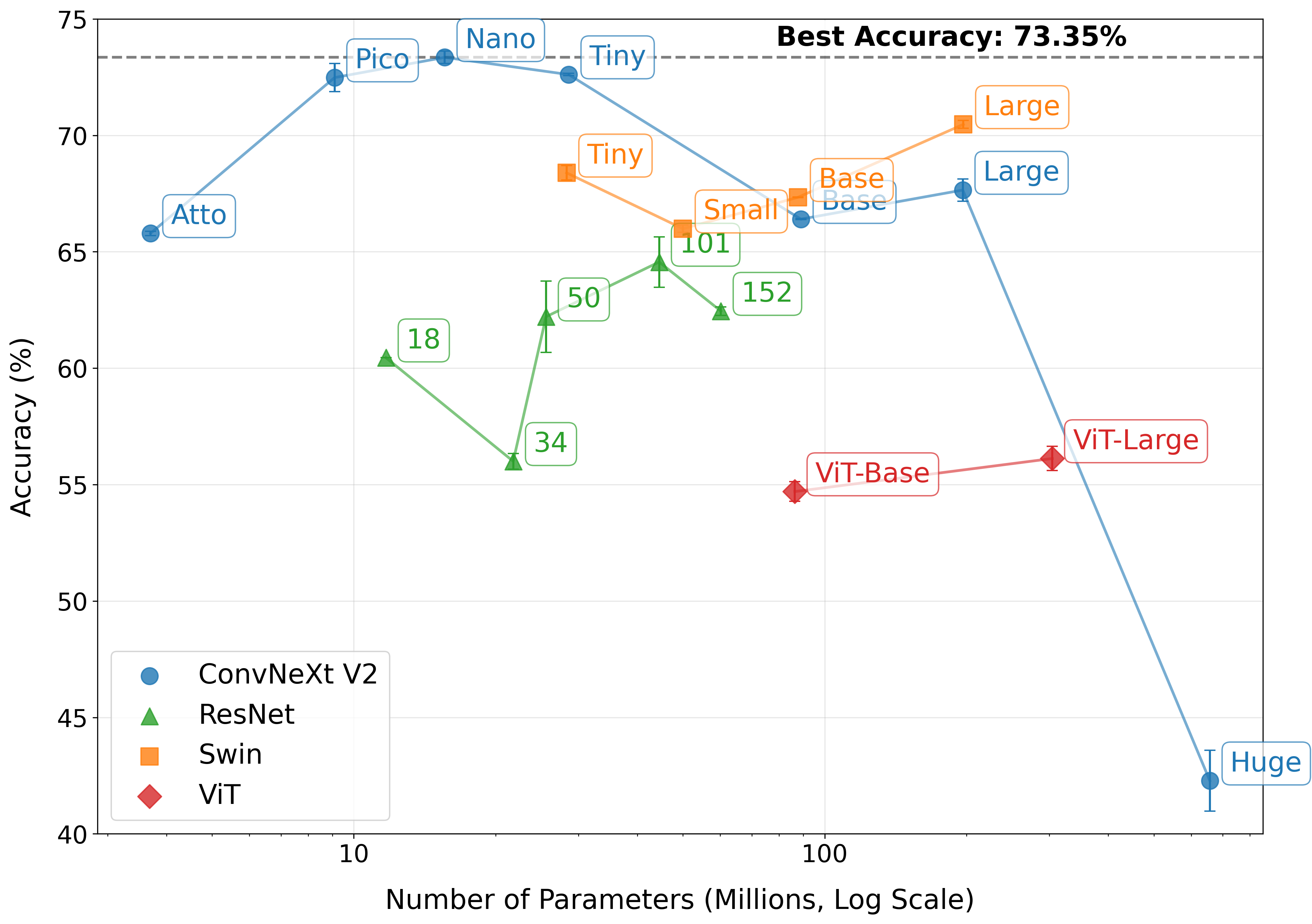}
    \caption{Linear classifier accuracy for calf identification across different 
    vision backbones on the 8-Calves dataset.}
    \label{fig:linear_classifier_accuracy}
\end{figure}
\begin{figure}[!htbp]
    \centering
    \includegraphics[width=0.95\textwidth]{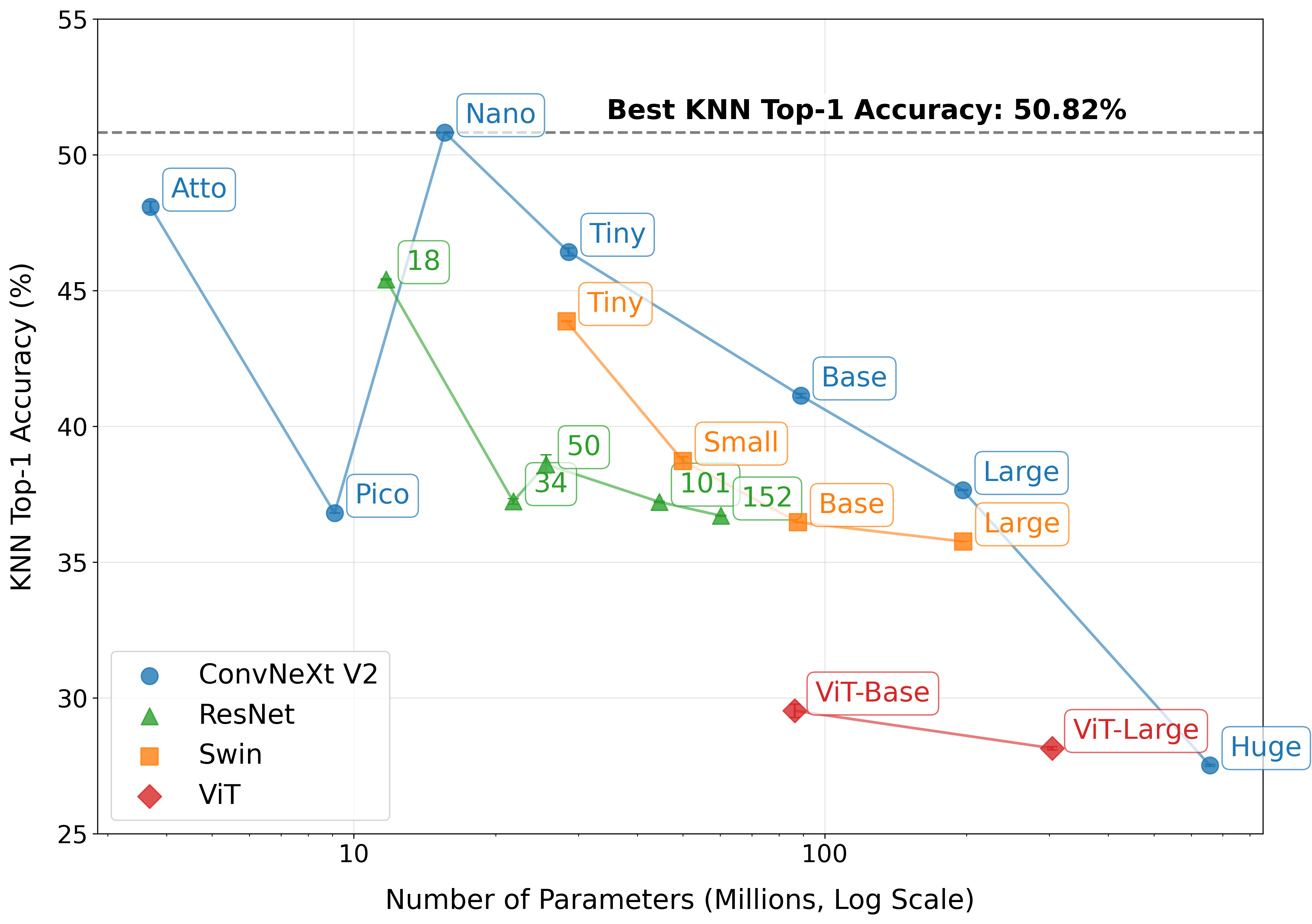}
    \caption{K-Nearest Neighbors (KNN) Top-1 retrieval accuracy 
    for calf identification on the 8-Calves dataset.}
    \label{fig:knn_top_1}
\end{figure}
\begin{figure}[!htbp]
    \centering
    \includegraphics[width=0.95\textwidth]{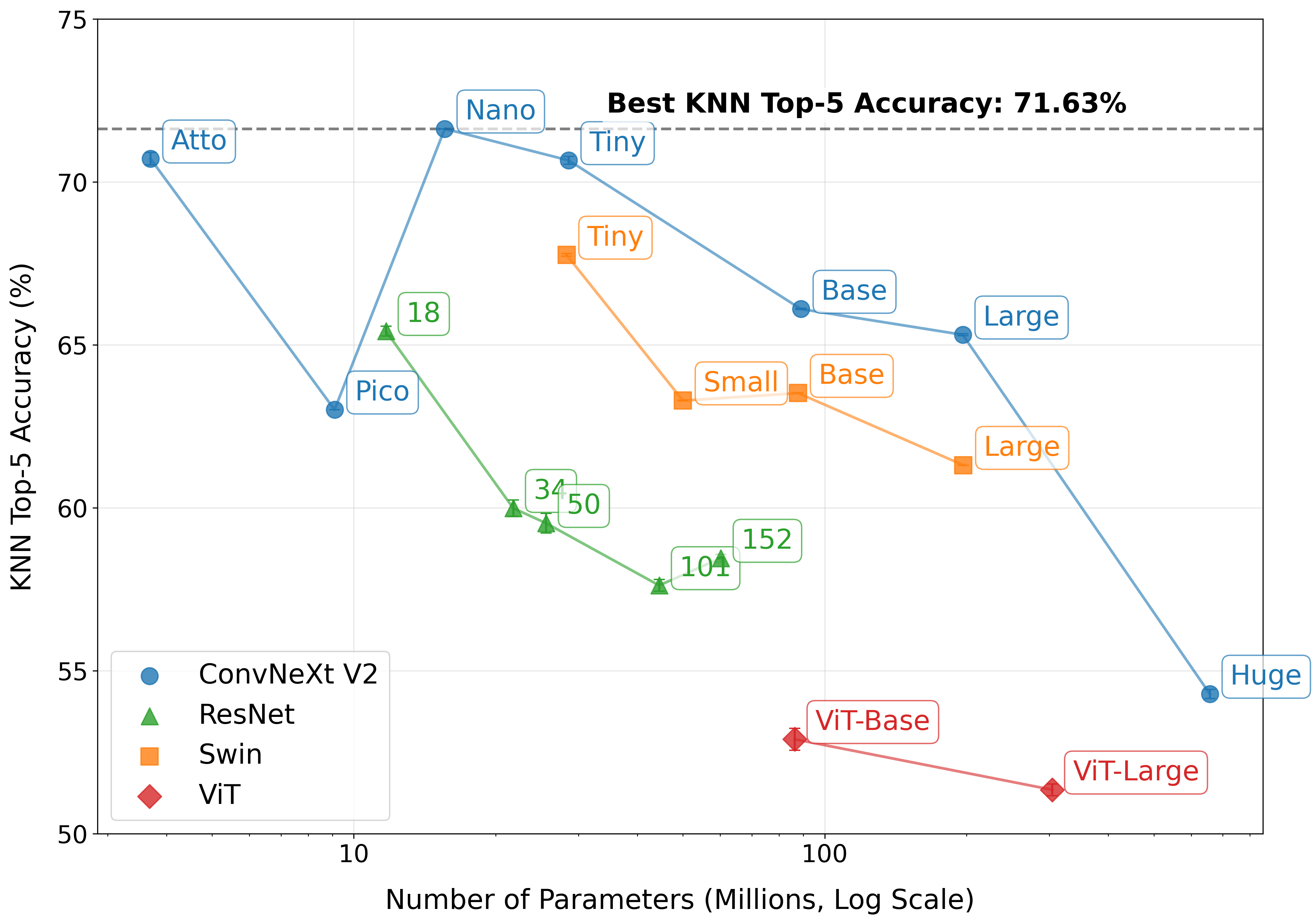}
    \caption{K-Nearest Neighbors (KNN) Top-5 retrieval accuracy 
    for calf identification on the 8-Calves dataset.}
    \label{fig:knn_top_5}
\end{figure}

\subsection{Results and Analysis}

As shown in Table ~\ref{tab:transfer_learning_model_family_comparison}, 
the ConvNextV2 Nano model (15.6M parameters) 
emerges as the top performer, achieving the highest accuracy (73.35\%) 
and KNN Top-1 retrieval rate (50.82\%). This result underscores the 
efficacy of smaller, optimized architectures for calf identification 
under significant occlusion. However, a severe overfitting issue is 
revealed as the model size increases. The performance of larger variants 
degrades catastrophically, with the Huge model (657.5M parameters) 
plummeting to an accuracy of just 42.28\%---a drop of 31 percentage points.

Scaling other architectures also yields diminishing returns, though less 
drastically, as shown in Table~\ref{tab:transfer_learning_model_family_comparison} 
and Figure~\ref{fig:linear_classifier_accuracy}. Swin Transformers see modest gains, 
with the Swin-Large model (70.47\% accuracy) being \(6.9\times\) larger than 
the Swin-Tiny variant for only a 3.8\% absolute improvement. Similarly, 
ResNet101 requires 3.8x more parameters than ResNet18 for a mere 4.1\% 
accuracy gain. Most strikingly, pure ViTs scale poorly across the board: 
ViT-Large barely surpasses ViT-Base (a 1.4\% difference) despite a \(3.5\times\) 
parameter increase. 

Our analysis reveals a clear trade-off: scaling up model size generally improves 
classification accuracy but compromises retrieval performance, as evidenced 
by dropping KNN Top-1/Top-5 rates in Figures~\ref{fig:knn_top_1} 
and~\ref{fig:knn_top_5}. This indicates that smaller 
models excel at capturing the fine-local details necessary for retrieval 
tasks, whereas larger models shift their focus towards global contextual 
patterns optimal for classification.

At the Base size level, ConvNextV2 and Swin Transformers deliver 
comparable performance, yet both significantly outperform Vision 
Transformers (ViT) despite utilizing identical pretraining data 
and methods. This result underscores the limitations of a pure 
transformer backbone (ViT) when compared to hybrid architectures 
that integrate convolutional inductive biases (Swin, ConvNextV2). 
Furthermore, the performance gap between the ResNet family and 
the leading models (ConvNextV2, Swin) can likely be attributed 
to the pretraining data discrepancy, as ResNet models were pretrained 
only on ImageNet-1K.
\begin{table}[!htbp]
\centering
\caption{Calf identification performance of ViT-Base under different pre-training strategies. The best-performing model is highlighted in red.}
\vspace*{10pt} % 10pt vertical space after the caption
\label{tab:vit_based_transformer_comparison}
\begin{tabular}{ccccc}
\toprule
\makecell{\textbf{Model}} & \makecell{\textbf{Accuracy} \\ \textbf{(\%)}} & \makecell{\textbf{KNN Top-1} \\ \textbf{(\%)}} & \makecell{\textbf{KNN Top-5} \\ \textbf{(\%)}} & \makecell{\textbf{Parameters}} \\
\midrule
DINOv2, Patch Size=14 & 66.14 \(\pm\) 1.10 & 41.75 \(\pm\) 0.00 & 62.78 \(\pm\) 0.00 & 86.6M \\
\textbf{BEiT, Patch Size=16} & \textcolor{red}{\textbf{67.61 \(\pm\) 0.90}} & \textcolor{red}{\textbf{42.40 \(\pm\) 0.00}} & \textcolor{red}{\textbf{63.33 \(\pm\) 0.00}} & 86.5M \\
DeiT, Patch Size=16 & 66.22 \(\pm\) 0.11 & 39.95 \(\pm\) 0.18 & 62.59 \(\pm\) 0.50 & 86.6M \\
CLIP, Patch Size=16 & 54.32 \(\pm\) 0.13 & 30.64 \(\pm\) 0.02 & 54.12 \(\pm\) 0.11 & 86.2M \\
ViT, Patch Size=16 & 54.70 \(\pm\) 0.42 & 29.53 \(\pm\) 0.25 & 52.90 \(\pm\) 0.34 & 86.4M \\
MAE, Patch Size=16 & 51.38 \(\pm\) 0.17 & 28.11 \(\pm\) 0.11 & 54.48 \(\pm\) 0.24 & 86.1M \\
\bottomrule
\end{tabular}
\end{table}

As shown in Table~\ref{tab:vit_based_transformer_comparison}, 
DeiT achieves competitive accuracy (66.22\%) but exhibits 
weaker local embedding cohesion, whereas CLIP and MAE deliver 
substantially poorer performance. This contrast underscores 
a critical misalignment between their pretraining objectives---multimodal 
alignment and pixel-level reconstruction, respectively---and the 
semantic feature demands of our task. Overall, the results 
demonstrate that pretraining strategies explicitly designed 
for semantic representation learning (e.g., BEiT's masked image 
modeling) yield superior transfer performance compared to those 
relying primarily on large-scale data (DINOv2) or low-level 
reconstruction objectives (MAE).

%% file: 5_Tracking/track.tex
\subsection{Experimental Setup}

To further establish the 8-Calves dataset as a comprehensive 
benchmark for Multi-Object Tracking (MOT), we evaluated two 
leading tracking algorithms, Bot-SORT 
\cite{aharon2022botsortrobustassociationsmultipedestrian} 
and ByteTrack \cite{bytetrack}. For a fair and consistent comparison, 
both trackers were integrated with the same fine-tuned YOLOv8m detector 
(described in Section 3), using the Ultralytics \cite{yolo11_ultralytics} implementations 
with their default parameters. It is noteworthy that our 
evaluation pipeline is fully deterministic, resulting 
in zero standard deviation across all reported metrics 
and ensuring the reproducibility of our results.

\subsection{Evaluation Metrics}

We report five standard MOT metrics to provide a holistic view of 
tracker performance:
\begin{itemize}
    \item MOTA (Multiple Object Tracking Accuracy) \cite{tang2019cityflowcityscalebenchmarkmultitarget}: 
    A composite score (ranging from $-\infty$ to 1) that consolidates 
    errors from false positives, false negatives, and identity switches. 
    Higher is better.
    \item MOTP (Multiple Object Tracking Precision) \cite{tang2019cityflowcityscalebenchmarkmultitarget}: 
    the misalignment between annotated and predicted object locations.
    Lower is better, indicating more precise localization.
    \item IDF1 Score: The ratio of correctly identified detections 
    to the average number of ground-truth and computed detections. 
    Higher is better, as it directly measures identity 
    preservation accuracy.
    \item Identity Switches: The total number of times a tracked 
    trajectory incorrectly changes its assigned ground-truth 
    identity. Lower is better.
    \item Track Fragmentations: The count of times a ground-truth 
    trajectory is interrupted. Lower is better.
\end{itemize}

These metrics collectively decouple overall detection 
and localization performance (MOTA, MOTP) from the challenging 
task of consistent identity preservation (IDF1, Identity 
Switches, Fragmentations).

\subsection{Results and Analysis}

Table~\ref{tab:tracking_results} presents the comprehensive 
multi-object tracking results. While both trackers achieve remarkably high MOTA scores (ByteTrack: 
0.927, Bot-SORT: 0.922) and low MOTP scores, confirming strong 
performance in detection coverage and precise localization, 
a different story emerges when examining identity preservation. 
The IDF1 scores for both trackers are notably low (approximately 0.27), 
and they are accompanied by a high number of Identity Switches 
(ID Switches > 420) and Track Fragmentations (> 2900). This 
divergence reveals a critical insight: our dataset successfully 
decouples the challenges of detection from those of 
identity-consistent tracking.

The high MOTA and low MOTP demonstrate that the task of 
simply detecting and precisely locating objects is manageable 
with modern detectors and trackers. However, the poor IDF1 
and high ID Switches underscore the extreme difficulty of 
maintaining correct identities through the frequent and 
severe occlusions. These results collectively underscore 
that while detection is highly accurate, preserving identity 
in such scenarios remains a primary challenge for current 
multi-object tracking paradigms.

\begin{table}[h]
\centering
\caption{Multi-object Tracking Accuracy (MOTA), Multi-Object Tracking Precision (MOTP) 
\cite{tang2019cityflowcityscalebenchmarkmultitarget} and Identity F1 (IDF1) 
of different trackers on the 8-Calves dataset.}
\label{tab:tracking_results}
\begin{tabular}{lccccc}
\toprule
\textbf{Tracker} & \textbf{MOTA Score} & \textbf{MOTP Score} & \textbf{IDF1 Score} & \makecell{\textbf{Identity} \\ \textbf{Switches}}  & \makecell{\textbf{Track} \\ \textbf{Fragmentations}}  \\
\midrule
ByteTrack & $0.92701$ & $0.18521$ & $0.26732$ & $442$ & $2900$\\
BotSort & $0.92194$ & $0.19410$ & $0.26918$ & $424$ & $3104$\\
\bottomrule
\end{tabular}
\end{table}

%% file: 6_Related_Work/related_work.tex
Our work builds upon and distinguishes itself from a body of existing datasets 
in animal monitoring, multi-object tracking, and general computer vision. 
The 8-Calves dataset is positioned to address the specific gap in the current 
landscape, particularly for vision tasks involving temporally persistent 
identities in natural, occlusion-rich environments.

The most directly comparable benchmark is 3D-POP \cite{3DPOP}, which tracks a
fixed group of pigeons in a controlled setting. Both datasets share 
three critical attributes: a consistent number of individuals, temporal 
continuity across recordings, and an emphasis on natural, ground-based 
motion. However, a fundamental distinction lies in the visual 
discernibility of the subjects. The subtle and homogeneous coat 
patterns of pigeons in 3D-POP \cite{3DPOP} limit its utility for appearance-based 
re-identification, whereas the distinct markings of Holstein Friesian 
calves in our dataset provide a robust foundation for evaluating 
models for identity preservation. Furthermore, 3D-POP provides shorter temporal windows, 
with videos lasting approximately 10 minutes, in contrast to our continuous 
hour-long recordings which are better suited for evaluating long-term 
tracking and behavior analysis.

Another highly relevant benchmark is the multimodal \textit{mmcows} dataset \cite{mmcows}, 
which presents significant challenges with 16 cows in a barn from 
a complex, occlusion-rich side-view, recorded from four cameras.
Our 8-Calves dataset offers a different benchmark focusing on the 
top-down perspective. This viewpoint is strategically chosen as 
prior work \cite{ANDREW2021106133,Andrew1,GaoActiveLearning} has 
demonstrated that the distinctive back patterns 
of cattle, clearly visible from above, enable highly accurate 
individual identification. This approach also offers a deployment 
advantage, requiring only a single overhead camera rather than 
a multi-camera system. Within this setting, we introduce severe 
challenges—including frequent group occlusions, motion blur, and 
varied poses during feeding—to evaluate the limits of top-down 
systems under realistic conditions. Furthermore, whereas \textit{mmcows} 
leverages multimodal data with sparse annotations 
(sampled every 15 seconds), our dataset provides a focused, 
vision-centric benchmark with high-frame-rate temporal continuity, 
which is essential for developing smooth, continuous tracking 
models.

Other cattle-specific datasets cater to different, often simpler, 
scenarios. CattleEyeView \cite{ong2023cattleeyeviewmultitasktopdownview} utilizes a fisheye lens in a constrained 
passage, artificially limiting animal poses and group dynamics. 
The Cattle Visual Behaviors (CVB) dataset \cite{zia2023cvbvideodatasetcattle} features a uniformly 
brown breed, making individual identification based on coat patterns 
impossible, and its open-field setting presents challenges distinct 
from our cluttered indoor environment. Cows2021 \cite{gao2021selfsupervisionvideoidentificationindividual} primarily contains 
videos of single animals passing through the frame, reducing 
multi-object tracking to a trivial task and does not represent 
the multi-animal, occlusion-rich setting we address.

Beyond the agricultural domain, DanceTrack \cite{DanceTrack} shares 
superficial similarities, such as a fixed camera and multi-object 
sequences. However, it diverges critically in its domain-specific 
constraints: individuals often wear identical costumes or obscure 
their faces, eliminating unique visual identifiers. Its sequences 
are short and choreographed, contrasting with the long-form, unstructured 
natural behaviors that define our dataset.

Foundational image datasets like MNIST \cite{MNIST} and CIFAR-10 \cite{CIFAR10} 
lack temporal dynamics and real-world complexities entirely. 
Sports-oriented benchmarks such as SportsMOT \cite{SportsMOT} and SoccerNet 
\cite{SoccerNET} focus on players in uniform, often with dynamic cameras, 
omitting the challenge of distinguishing unique individuals 
in a fixed context. General-purpose object detection datasets 
like ImageNet \cite{ImageNet}, MS-COCO \cite{MSCOCO}, and PASCAL VOC \cite{pascal_voc} prioritize 
category-level diversity over temporal continuity or instance-level 
identity preservation. While they include occlusions, these occur sporadically 
rather than systematically, as in closed-group settings.

By integrating temporal richness, visually distinct identities, 
and controlled yet realistic group dynamics, the 8-Calves dataset 
fills a unique niche. It provides a benchmark for vision tasks—detection, tracking, and 
re-identification—within a context directly relevant to 
precision livestock farming. Unlike 3D-POP \cite{3DPOP}, 
it provides longer sequences with visually unambiguous patterns; 
unlike \textit{mmcows}, it offers a focused, high-frame-rate vision benchmark 
from a top-down angle, without relying on multimodal data;
unlike DanceTrack \cite{DanceTrack}, it avoids irrelevant domain-specific
biases, focusing instead on natural, unscripted motion; and compared 
to general or static datasets, it systematically 
introduces the challenges of occlusion and identity consistency over 
time, thereby enabling evaluation of models for automated 
animal monitoring.

%% file: 7_Conclusion/conclusion.tex
In this paper, we introduced the 8-Calves dataset, a novel 
benchmark designed to address the challenges of multi-animal 
detection, tracking, and identification in a realistic, 
occlusion-heavy agricultural environment. Through a semi-automated 
pipeline combining a fine-tuned detector and a state-of-the-art 
tracker followed by meticulous manual validation, we provided 
a high-quality dataset comprising over 537,000 bounding boxes 
with temporal identity labels.

Our comprehensive benchmarking demonstrates the utility and 
distinctiveness of this dataset. In object detection, while 
modern architectures achieve near-perfect performance under 
a lenient IoU threshold (mAP50), their performance significantly 
diverges on more stringent metrics (mAP50:95 and mAP75), 
highlighting the challenge of precise localization posed by 
our data. The observed overfitting in certain model families 
further underscores the dataset's value for model selection 
and evaluation.

In calf identification, our experiments reveal a 
trade-off: while scaling model size can benefit classification 
accuracy, it often compromises the local feature cohesion 
essential for retrieval tasks. We identified that smaller, 
efficiently designed architectures like ConvNextV2 Nano 
achieve an optimal balance, delivering superior accuracy 
and retrieval performance. Furthermore, we established 
that pretraining strategies focused on semantic representation 
learning (e.g., BEiT) yield better transferability to our task 
than those reliant on massive data or low-level reconstruction.

Finally, our multi-object tracking benchmarks demonstrate 
that while detection is highly accurate (MOTA > 0.92), current 
methods are insufficient for maintaining identity through 
severe occlusions (IDF1 $\approx$ 0.27). This stark contrast 
underscores the value of our dataset as a suitable benchmark 
for developing and evaluating more robust, 
identity-consistent tracking algorithms.

In conclusion, we present the 8-Calves dataset as a benchmark designed 
to be useful for developing computer vision models that meet the 
practical demands of livestock management. The tracking and 
identification capabilities evaluated in this work are directly 
relevant to automating individual-animal management. Potential 
practical applications include:

\begin{itemize}
    \item Enhance welfare monitoring through continuous, non-invasive 
    assessment of behavior and locomotion.
    \item Improve operational efficiency by reducing the labor 
    required for manual observation and enabling data-driven decisions.
    \item Boost economic sustainability by enabling early disease 
    detection, which can reduce treatment costs and improve growth rates. 
\end{itemize}

Looking forward, this benchmark opens up two promising research directions: first, extending 
the dataset with textual descriptions of visual features to 
enable multimodal models that can not only identify but also 
describe individual calves in natural language, paving the 
way for intelligent human-assistive systems; second, given 
the labor-intensive annotation process, prioritizing the 
development of efficient self-supervised methods that can 
learn robust representations directly from unlabeled video, 
thereby overcoming the primary bottleneck in model development. 
By addressing these challenges, the 8-Calves dataset aims to 
contribute to innovation in automated, non-invasive animal 
monitoring and thereby support the advancement of precision 
livestock farming.

%% file: 8_Appendix/appendix.tex
\section*{Appendix A}
\begin{table}[!htbp]
\centering
\setlength{\tabcolsep}{6pt}
\small
\caption{Hyperparameter selection for fine-tuning different Object Detection models.}
\vspace*{10pt} % 10pt vertical space after the caption
\begin{tabularx}{\textwidth}{|l|*{4}{>{\raggedright\arraybackslash}X|}}
\hline
%\makecell{\textbf{horizontal} \\ \textbf{flip}}
\textbf{Hyperparameter} & \textbf{Conditional DETR} & \textbf{Facebook DETR} & \textbf{YOLOS \newline Small} & \textbf{YOLO Detectors} \\
\hline
Image Size & 640 & 640 & 640 & 640 \\
\hline
Epochs & 100 & 100 & 100 & 100 \\
\hline
Optimiser & AdamW \newline Weight Decay: 1e-4 & AdamW \newline Weight Decay: 1e-4 & AdamW \newline Weight Decay: 1e-4 & SGD \newline Momentum: 0.937 \newline Weight Decay: 5e-4 \\
\hline
Learning Rates & backbone: 1e-5 \newline head: 1e-4 & backbone: 1e-5 \newline head: 1e-4 & entire model: 2.5e-5 & initial: 0.01 \newline final: 0.0001 \\
\hline
LR Scheduler & constant & constant & cosine & linear \\
\hline
Batch Size & 5 & 5 & 5 & 5 \\
\hline
Gradient Clipping & 0.1 & 0.1 & 0.0 & 0.0 \\
\hline
Warmup & 0 steps & 0 steps & 0 steps & 3 epochs \newline (momentum: 0.8, \newline bias LR: 0.1) \\
\hline
Early Stopping Patience & 10 epochs & 10 epochs & 10 epochs & 10 epochs \\
\hline
Number of Classes & 2 \newline (cow and background) & 2 \newline (cow and background) & 2 \newline (cow and background) & 1 \\
\hline
\end{tabularx}
\label{tab:hyperparams}
\end{table}
% \newpage
% \input{7_Appendix/neurips_checklist}